\documentclass{article}

\usepackage{microtype}
\usepackage{graphicx}
\usepackage{booktabs} 
\usepackage{subfig}
\usepackage{relsize} 
\usepackage{hyperref}

\usepackage[accepted]{icml2020}

\usepackage{amsmath}
\usepackage{ltablex}
\usepackage{longtable}
\usepackage{siunitx}
\usepackage{todonotes}
\usepackage{comment}
\usepackage{booktabs}       
\usepackage{amsfonts}       
\usepackage{nicefrac}       
\usepackage{microtype}      
\graphicspath{{figures/}}
\usepackage{multirow}
\usepackage{xcolor}
\newcommand\trapsnet{{\textsc{TraPSNet}}}
\newcommand\symnet{{\textsc{SymNet}}}
\newcommand\torpido{{\textsc{ToRPIDo }}}

\newcommand\A{{\cal A}}
\newcommand\K{{\cal K}}
\newcommand\tta{{\bf a}}
\newcommand\T{{\cal T}}
\newcommand\SP{{\cal SP}}
\newcommand\R{{\cal R}}
\newcommand\C{{\cal C}}

\newcommand\F{{\cal F}}
\newcommand\NF{{\cal NF}}

\makeatletter
\def\old@comma{,}
\catcode`\,=13
\def,{%
  \ifmmode%
    \old@comma\discretionary{}{}{}%
  \else%
    \old@comma%
  \fi%
}
\makeatother

\mathchardef\mhyphen="2D 

\icmltitlerunning{Symbolic Network: Generalized Neural Policies for Relational MDPs}

\begin{document}
\twocolumn[
\icmltitle{Symbolic Network: Generalized Neural Policies for Relational MDPs}

\icmlsetsymbol{equal}{*}

\begin{icmlauthorlist}
\icmlauthor{Sankalp Garg}{iit}
\icmlauthor{Aniket Bajpai}{iit}
\icmlauthor{Mausam}{iit}
\end{icmlauthorlist}

\icmlaffiliation{iit}{Indian Institute of Technology Delhi}

\icmlcorrespondingauthor{Sankalp Garg}{sankalp2621998@gmail.com}
\icmlcorrespondingauthor{Aniket Bajapi}{quantum.computing96@gmail.com}
\icmlcorrespondingauthor{Mausam}{mausam@cse.iitd.ac.in}

\icmlkeywords{Machine Learning, RMDP}

\vskip 0.3in  
]

\printAffiliationsAndNotice{}
\begin{abstract}



A Relational Markov Decision Process (RMDP) is a first-order representation to express all instances of a single probabilistic planning domain with possibly unbounded number of objects. Early work in RMDPs outputs generalized (instance-independent) first-order policies or value functions as a means to solve \emph{all} instances of a domain at once. Unfortunately, this line of work met with limited success due to inherent limitations of the representation space used in such policies or value functions. Can neural models provide the missing link by easily representing more complex generalized policies, thus making them effective on all instances of a given domain?

We present \symnet{}, the first neural approach for solving RMDPs that are expressed in the probabilistic planning language of RDDL. \symnet{} trains a set of shared parameters for an RDDL domain using training instances from that domain. For each instance, \symnet{} first converts it to an instance graph and then uses relational neural models to compute node embeddings. It then scores each ground action as a function over the first-order action symbols and node embeddings related to the action. Given a new test instance from the same domain, \symnet{} architecture with pre-trained parameters scores each ground action and chooses the best action.  This can be accomplished in a single forward pass {\em without any retraining} on the test instance, thus implicitly representing a neural generalized policy for the whole domain. Our experiments on nine RDDL domains from IPPC demonstrate that \symnet{} policies are significantly better than random and sometimes even more effective than training a state-of-the-art deep reactive policy from scratch. 
\end{abstract}
\section{Introduction}
\label{introduction}
A Relational Markov Decision Process (RMDP) \cite{relational_mdp} is a first-order, predicate calculus-based representation for expressing instances of a probabilistic planning domain with a possibly unbounded number of objects. An RMDP {\em domain} has object types, relational state  predicate and action symbols that are applied over objects, 
first order transition templates that specify  probabilistic effects associated with action symbols, and a first-order reward structure. 
A domain {\em instance} additionally specifies a  set of objects and a start state, thus defining a ground MDP with a known start state \cite{kolobov-uai12}).
\emph{Relational} planners aim to produce a single {\em generalized} policy that can yield a ground policy for {\em all} instances of the domain, with little instance-specific computation. \emph{Domain-independent} planners are representation-specific, but domain-agnostic, making them applicable to all domains expressible in the language. In this paper, we design a domain-independent relational planner.

RMDP planners, in their vision, expect to scale to very large problem sizes
by exploiting the first-order structures of a domain -- thereby reducing the curse of dimensionality. 
Traditional RMDP planners attempted 
to find a generalized {\em first-order} value function or policy using symbolic dynamic programming \cite{relational_mdp}, or by approximating them via a function over first-order basis functions (e.g., \cite{guestrin-rmdp,sanner-boutilier}).  Unfortunately, these methods met with rather limited success, for e.g., no relational planner participated in International Probabilistic Planning Competition (IPPC)\footnote{\url{http://www.icaps-conference.org/index.php/Main/Competitions}} after 2006, even though all competition domains were relational. We believe that this lack of success may be due to the inherent limitations in the representation power of a basis function-based representation.  Through this work, we wish to revive the research thread on RMDPs and explore if neural models could be effective in representing these first-order functions.


We present \textbf{Sym}bolic \textbf{Net}Work (\symnet), the first domain-independent neural relational planner that computes generalized policies for RMDPs that are expressed in the symbolic representation language of RDDL \cite{rddl}. \symnet{} outputs its generalized policy via a neural model whose all parameters are specific to a domain, but tied among all instances of that domain. So, on a new test instance, the policy can be applied out of the box using pre-trained parameters, i.e., without any retraining on the test instance. \symnet{} is domain-independent because it converts an RDDL domain file (and instance files) completely automatically into neural architectures, without any human intervention.

\symnet{} architecture uses two key ideas. First, it visualizes each state of each domain instance as a graph, where nodes represent the \emph{object tuples} that are valid arguments to some relational predicate. An edge between two nodes indicates that an action causes predicates over these two nodes to interact in the instance. The values of predicates in a state act as features for corresponding nodes.  \symnet{} then learns node (and state) embeddings  for these graphs using graph neural networks. 
Second, \symnet{} learns a neural network to represent the policy and value function over this graph-structured state.   To learn these in an instance-independent way, we recognize that most ground actions are a first-order action symbol applied over some object tuple.  \symnet{} scores such ground actions as a function over the action symbol and the relevant embeddings of object tuples. After training all model parameters using reinforcement learning over training instances of a domain, \symnet{} architecture can be applied on any new (possibly larger) test problem without any further retraining.


We perform experiments on nine RDDL domains from IPPC 2014 \cite{ippc14}. Since no planner exists that can run without computation on a given instance,  we compare \symnet{} to random policies (lower bound) and  policies trained from scratch on the test instance. We find that \symnet{} obtains hugely better rewards than random, and is quite close to the policies trained from scratch -- it even outperforms them in $28\%$ instances. 
Overall, we believe that our work is a step forward for the difficult problem of domain-independent RMDP planning. We release the code of \symnet{} for future research.\footnote{\url{https://github.com/dair-iitd/symnet}}

\section{Background and Related Work}
\label{background}

\subsection{Probabilistic Planning}

\noindent
\textbf{Markov Decision Process (MDP): }
A (ground) finite-horizon MDP \cite{bellman57,puterman94} with a known start state is formalized as a tuple $<S, A, T, R, H, s_0, \gamma>$, where $S$ is the set of states, $A$ is the set of actions, $T$ is the transition model $S \times A \times S \rightarrow [0,1]$, 
$R$ is the reward model $S \times A \times S \rightarrow \mathbb{R}$, $H$ is the horizon and $s_0$ is the start state, and $\gamma$, the discount factor. 
Probabilistic planning problems are often expressed via a factored MDP \cite{mausam&kolobov12}. It factors a state $s$ into a set of state variables $X$, i.e., $s=\{x_i\}_{i=1}^{|X|}$. $T$ may also be factored, defined via, e.g., a DBN, dynamic Bayesian network \cite{dbn}, which maintains the conditional probability table $T^f$ of $x_i^{'}$ dependent on action $a$, previous state $s$, and lower valued $x_j^{'}$s, i.e., $T^f(x_i^{'}|s,a, x_1^{'}, \ldots x_{i-1}^{'})$. The joint probability $T(s,a,s^{'})$ $=\prod_{x_i^{'}\in s^{'}}T^f(x_i^{'}|s,a,x_1^{'},\ldots x_{i-1}^{'})$. In practice, these models are compact, and an $x_i^{'}$ depends only on a small number of other state variables. 

\noindent
\textbf{Relational Markov Decision Process (RMDP): }
An RMDP $<\C, \SP, \A, O, \T, \R, H, s_0, \gamma>$ is a first-order representation of a factored MDP \cite{relational_mdp}, expressed via objects, predicates and functions. 
Here, $\C$ is a set of classes (types), $\SP$ is the set of state predicate symbols. 
$\A$ is a set of action predicate symbols, 
$O$ represents a set of domain objects, each associated with single type from $\C$. It is a first order representation because different sets of objects $O$ can construct different ground MDPs.

Each predicate symbol is declared to take as argument a tuple of object types. A predicate symbol (action symbol) applied over a type-consistent tuple of object variables forms a state variable (respectively, ground action). A ground-state $s$ is, thus, a complete assignment of all predicate symbols $\SP$ applied on all type-consistent object tuples from $O$ (also denoted by $\SP_O$). Similarly, the set of all ground actions ($A$) can be defined as $\A_O$ -- all-action symbols applied on all type-consistent object tuples. We also denote the ground state space $S$ by $\mathcal P(\SP_O))$, where $\mathcal P$ denotes the powerset. 
Transition and reward models for an RMDP are defined at the schema level
through different languages, e.g., PPDDL \cite{ppddl} and, our focus,  RDDL \cite{rddl}. 

Research in Relational MDPs explores ways to represent and construct first-order (generalized) value functions or policies, which can be used directly on a new test instance. Example representations for these include regression trees \cite{mausam&weld03}, decision lists \cite{fern-jair}, extensions of algebraic decision diagrams \cite{joshi-jair11}, and linear combinations of basis functions \cite{guestrin-rmdp,sanner-uai05}. We believe a reason for limited success of RMDP algorithms is the inherent limitation of these representations. We study the use of deep neural models for representing such generalized functions. Moreover, to the best of our knowledge, we are the first to develop relational planners for domains  expressed in RDDL.

\noindent
\textbf{Relational Dynamic Decision Language (RDDL): }
RDDL 
has been the language of choice for the last three IPPCs. It divides 
$\SP$ into non-fluent ($\NF$) and fluent ($\F$) symbols. Non-fluents are the state variables that do not change with time in a given instance, but may be different across problem instances.  Fluents represent state variables that change with time (due to actions or natural dynamics). 
RDDL splits an RMDP into two separate files, one for the whole domain (that has types, predicates, transitions, and rewards), and the other for the instance (that has objects, non-fluent values, and fluent values for initial state).  RDDL uses additive rewards -- the total reward is the sum of local rewards collected for satisfying different properties in a state. It factors the transition function via an underlying DBN semantics. There exist algorithms that convert an RDDL instance into a ground DBN \cite{rddl}. 

\noindent
{\bf Running Example: } We use a simplified Wildfire domain as our example.  It has a grid where each cell may have fuel, causing it to burn. The goal is to have the least damage to the grid by either putting out the fire or cutting out the fuel supply. The DBN for the domain is show in Figure \ref{ModifiedWildfireDBN}.


There are two classes, $\mathcal{C} = \{x_{pos}, y_{pos}\}$: $x$ and $y$ coordinate of the grid cell. Domain has two fluent symbols $\F = \{ burning, out\mhyphen of\mhyphen fuel \}$, representing the current burning state and the fuel state of the cell. Both fluent symbols take a cell $(x,y)$ as its arguments.  The non-fluents represent costs and topology, $\NF = \{ CostTgtBurn, CostNTgtBurn, Neighbour, Target \}$. The non fluent symbol $Neighbour$ takes four arguments $(x, y, x', y')$, since it defines the topology of the grid. $Target$ has arguments $(x,y)$. 
$\A = \{ put\mhyphen out, cut\mhyphen out, finisher\}$. 
First two action symbols take arguments $(x, y)$ -- they put out fire and cut out fuel supply at a cell. There is one global action $finisher$,  which puts out fire in all the cells simultaneously. Reward (negative) in each time step adds $CostTgtBurn$ for each target cell that is burning and $CostNTgtBurn$ for each non-target cell that is burning. 
In a problem instance, say there are three objects ${O} = \{ x1,x2,y1 \}$. This implies a problem with two cells $(x1,y1)$, and $(x2,y1)$. Say the target cell is $(x1,y1)$ and that these are connected, i.e., $Neighbour(x1,y1,x2,y1)=1$.

{\bf RDDL vs PPDDL:} PPDDL and RDDL have significant differences in their modeling choices. PPDDL uses correlated effects, whereas RDDL naturally models parallel effects. Thus, RDDL handles no-op actions with underlying natural dynamics better. RDDL rewards are state-dependent and sum over all objects satisfying a property, whereas PPDDL has both goals and rewards, but its rewards are associated with action transitions and do not aggregate over objects.





 \subsection{Reinforcement Learning}
 Reinforcement Learning (RL) refers to planning problems without known transition and rewards, necessitating learning from experience. State of the art approaches for RL are neural, which approximate policy and value functions through deep neural models. We use the Asynchronous Advantage Actor-Critic (A3C) \cite{a3c} as our underlying RL algorithm. A3C uses two neural networks 1) $\theta_\pi$ to represent the policy (mapping from a state to distribution over actions) and 2) $\theta_V$ to estimate the state value (long term discounted reward starting in a state). 
 Policy parameters are optimized to prefer an action that increases a state's advantage function -- difference between its value when taking that action and its overall value. Value parameters optimize the MSE loss between observed and predicted long-term rewards.


\subsection{Graph Neural Networks} 
Graph Neural Networks input a graph and learn latent space embeddings for each node, based on the its individual features and the local connectivity structure. Examples include Graph Convolution Networks (GCN) \cite{kipf-iclr17}, and Graph Attention Networks (GAT) \cite{gat}.
We use GAT, 
 which computes a node embedding by using a weighted attention for each of neighbouring nodes. Specifically output node embedding 
$     \overline{v}_i{'} = \sigma (\frac{1}{K}\sum_{k=1}^K \sum_{j\in N_i} \alpha_{ij}^{k} \mathbf{W}^k \overline{v}_j)$, 
 where $\overline{v}_i$ is the input feature of node $v_i$, $N_i$ is its neighbours, $\mathbf{W}^k$ is a trainable weight matrix, $k$ is the multi-head hyperparameter and $\alpha_{ij}^k = softmax_j(f(\mathbf{W}^k \overline{v}_i,\mathbf{W}^k \overline{v}_j))$ is the normalized self attention coefficient for any non-linear function ($f$), which in our implementation is LeakyReLU \cite{leaky-relu}.

\subsection{Transfer Learning for Probabilistic Planning}

There having been several classical \cite{taylor-stone,valuetransferclassical,policytransferclassical} and neural \cite{actor-mimic,tscl,deep-symbolic-rl,darla} approaches for transfer learning in RL. Recent work studies transfer learning for symbolic planning problems, e.g., Groshev et al. (\citeyear{abbeel18}) 
for deterministic planning problems. 
ASNets, Action-Schema Networks \cite{asnet18,shen2019guiding}, tackle a problem similar to ours but for goal-oriented subset of PPDDL. 
While an RDDL instance can be converted automatically into propositional PPDDL, an RDDL domain cannot always be converted into relational PPDDL -- hence we cannot directly compare against ASNets. 
Issakkimuthu et al. (\citeyear{fern18}) devise a neural framework to learn a policy for (ground) RDDL MDPs from scratch. Their constraint on non-transferability is due to the fixed size of fully connected layers in the neural network. \torpido{} achieves transfer across RDDL problem instances of the same domain \cite{torpido}; it can only transfer over {\em equi-sized} problems due to its fixed size action decoder. 


Our previous work \trapsnet{} is closest to \symnet{}, as it can transfer to different-sized instances of an RMDP \cite{trapsnet}. It constructs a graph, which uses single object as nodes, and non-fluent based edge. It encodes each node in embedding space and computes the score for a ground action based on the applied action template, and object embedding. However, \trapsnet{} makes the restrictive assumptions that the domains have exactly one binary non-fluent, and all the rest are unary fluents or non-fluents,  
and that each action symbol is parameterized by \emph{exactly} one object. These assumptions do not hold in several RDDL domains. 




\section{Problem Formulation}
\label{formulation}
Given an RMDP domain $D=<\C, \SP, \A, \T, \R, H, \gamma>$ expressed in RDDL, we wish to learn a generalized policy $\pi^{D}$, which can be applied to all instances of $D$ and maximizes the discounted sum of expected rewards over a finite horizon $H$. Given a test problem instance $I_{t}=<O, s_{0}>$ from $D$, this generalized policy can yield an instance-specific policy $\pi^{D}(I_{t}): \mathcal P(\SP_O) \rightarrow \A_O$, without any training on $I_{t}$. 
The RMDP learning problem can be seen in terms of multi-task learning over several problem instances in $D$: 
given $N$ randomly selected problem instances $I_1$, $I_2$, ..., $I_N$ (possibly of different sizes) from $D$, we wish to learn the weights $\phi$ of a neural network, such that $\pi^{D}(I_i; \phi)$ is a good (high-reward) policy for problem instance $I_i$. A good generalized policy is one which, without training, achieves high reward values on the new instance $I_{t}$. 



\section{The \symnet{} Framework}
\label{framework}

\begin{figure*}[t]
    \centering
    \includegraphics[width=\textwidth]{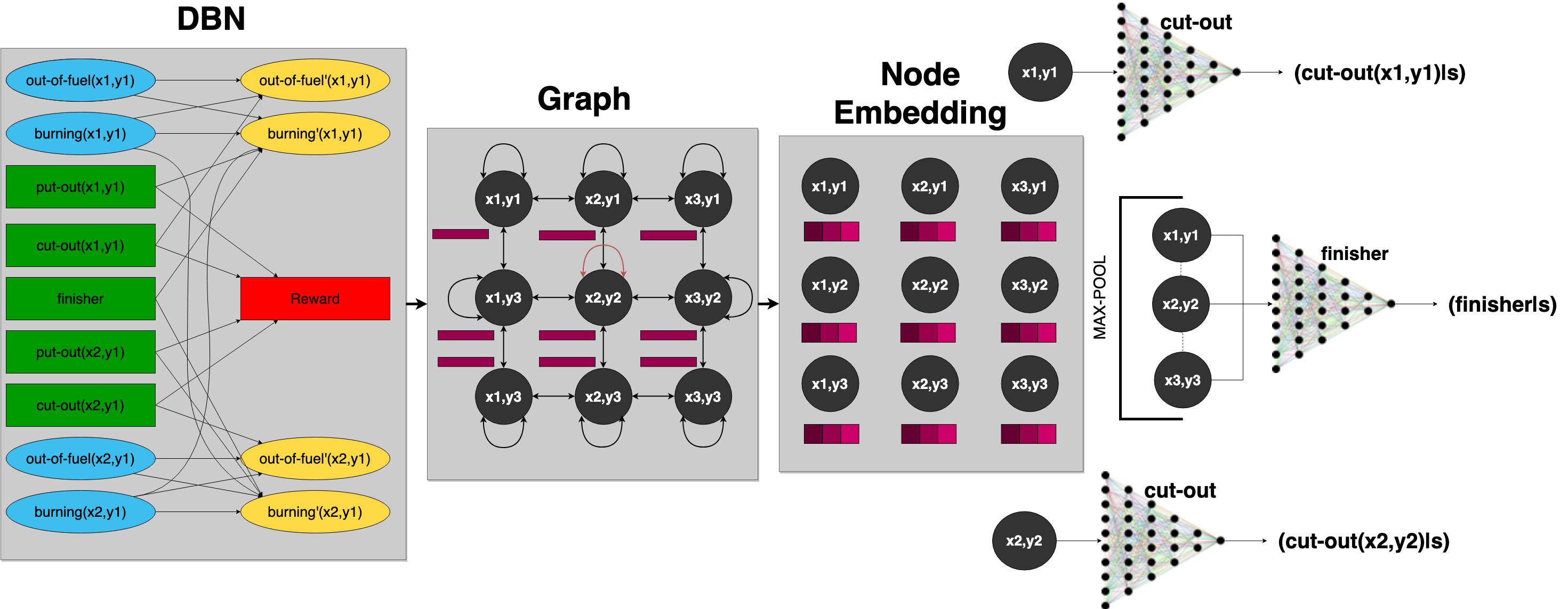}
    \vspace*{-2ex}
    \caption{Policy network for \symnet{} demonstrated on $2\times1$ wildfire domain. Fully Connected Network is used in Action Decoder.}
    \label{network}
\end{figure*}

We now present \symnet{}'s architecture for training a generalized policy for a given RMDP domain. We follow existing research to hypothesize that for any instance of a domain, we can learn a representation of the current state in a latent space and then output a policy in latent space, which is decoded into a ground action. To achieve this, \symnet{} uses three modules: (1) problem representation, which constructs an instance graph for every problem instance, (2) representation learning, which learns embeddings for every node in the instance graph, and for the state, and (3) policy decoder, which computes a value for every ground action, outputting a mixed policy for a given state.  All parameters of representation learning and policy learning modules are shared across all instances of a domain. \symnet's full architecture is shown in Figure \ref{network}.



\subsection{Problem Representation}

We follow \trapsnet{}, in that we continue the general idea of converting an instance into an instance graph and then learning a graph encoder to handle different-sized domains. However, the main challenge for a general RMDP, one that does not satisfy the restricted assumptions of \trapsnet{}, is in defining a coherent graph structure for an instance.
The first key question is what should be a node in the instance graph. \trapsnet{}'s approach was to use a single object as nodes, as all fluents (and actions) in its domains took single objects as arguments. This may not work for a general RMDP since it's fluents and actions may take several objects as arguments. Secondly, how should edges be defined. Edges represent the interaction between nodes. \trapsnet{} defined them based on the one binary non-fluent in its domain. A general RMDP may not have any non-fluent symbol or may have many (possibly higher-order) non-fluents.



Last but not least, the real domain-independence for \symnet{} can be achieved only when it parses an RDDL domain file without any human intervention. 
This leads to a novel challenge of reconciling multiple different ways in RDDL to express the same domain. In our running example, connectivity structure between cells may be defined using non-fluents $y\mhyphen neighbour(y, y')$, $x\mhyphen neighbour(x, x')$, or using a quarternary non-fluent $neighbour(x,y,x,y')$. Since both these representations represent the same problem, an ideal desideratum is that the graph construction algorithm leads to the same instance graph in both cases. But, this is a challenge since the corresponding RDDL domains may look very different. 
While, in general, this problem seems too hard to solve, since it is trying to judge logical equivalence of two domains, \symnet{} attempts to achieve the same instance graphs in case the equivalence is within non-fluents. 

\begin{figure}[t]
    \centering
    \includegraphics[width=\columnwidth]{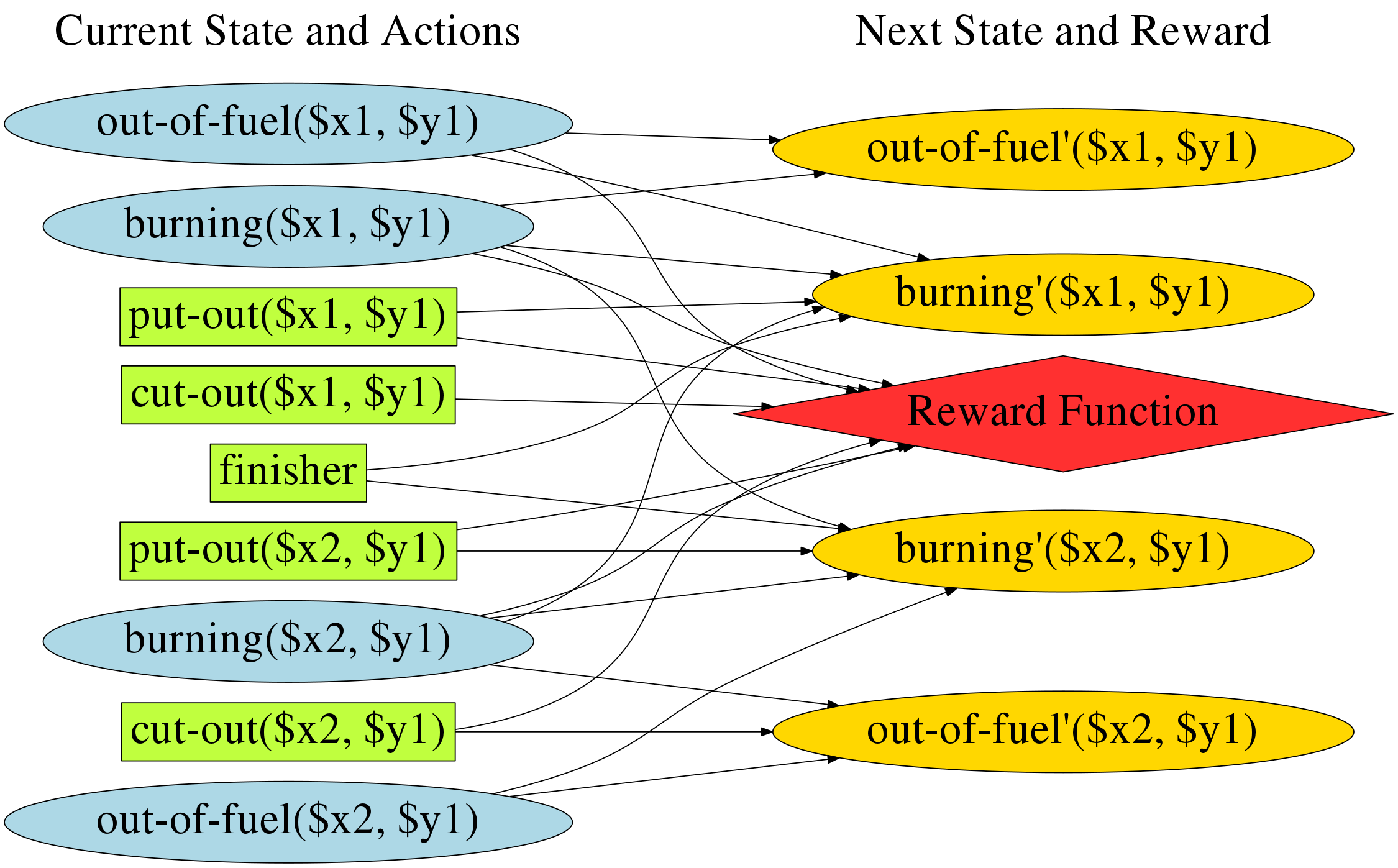}
    
    \caption{DBN for a modified wildfire problem.}
    \label{ModifiedWildfireDBN}
\end{figure}

To solve these problems, we make the observation that dynamics of an RDDL instance ultimately compile to a ground DBN with nodes as state variables (fluent symbols applied on object tuples) and actions (action symbols applied on object tuples).\footnote{done automatically using code from \url{https://github.com/ssanner/rddlsim}} DBN exposes a connectivity structure that determines which state variables and actions directly affect another state variable. It additionally has conditional probability tables (CPTs) for each transition. Figure \ref{ModifiedWildfireDBN} shows an example of a DBN for our running example instance. Here, left column is for current time step, and right for the next one.
The edges represent which state and action variables affect the next state-variable. We note that the ground DBN does not expose non-fluents since its values are fixed, and their dependence can be compiled directly into CPTs. 

\symnet{} converts a ground DBN to an instance graph. 
It constructs a node for every unique \emph{object tuple} that appears as an argument in any state variable in the DBN. Moreover, two nodes are connected if the state variables associated with two nodes influence each other in the DBN through some action. This satisfies all our challenges. First, it goes beyond an object as a node, but only defines those nodes that are likely important in the instance. Second, it defines a clear semantics of edges, while maintaining its intuition of ``directly influences.'' Finally, it can handles some variety of non-fluent representations for the same domain. Since the DBN does not even expose non-fluent state variables, and compiles them away, 
same instancs encoded with different non-fluent representations often yield yield same ground DBNs and thus the same instance graphs. 

\noindent{\bf Construction of Instance Graph:}
We now formally describe the conversion of a DBN into a directed instance graph, 
$G = (V,E)$, where $V$ is the set of vertices and $E$ is the set of edges.
$G$ is composed of $\K=|\A|+1$ disjoint subgraphs $G_j=(V_j, E_j)$. Intuitively, each graph $G_j$ has information about influence of each individual action symbol $\tta_j\in\A$. $G_\K$ represents the influence of the full set $\A$, and also the natural dynamics. In our example $\K=4$ since we have three action symbols: put-out, cut-out and finisher.

To describe the formal process, we define three analogous sets: $O_f$, $O_{nf}$ and $O_a$. $O_f$ represents the set of all object tuples that act as a valid argument for any fluent symbol. $O_{nf}$ and $O_a$ are analogous sets for non-fluent and action symbols. In our running example, $O_f = \{ (x1,y1), (x2,y1) \}$, $O_{nf} = \{ (x1,y1), (x2,y1), (x1,y1,x2,y1), (x2,y1,x1,y1)\}$, and $O_a = \{ (x1,y1), (x2,y1) \}$. Nodes in the instance graph associate with object tuples. We use $o_v$ to denote the object tuple associated with node $v$.
\symnet{} converts a DBN into an instance graph as follows:
\begin{enumerate}
\vspace*{-1ex}
    \item The distinct object tuples in fluents form the nodes of the graph, i.e. $V_j = \{v | o_v \in O_f\}, \forall j$. For the example, each $V_j =$ different copies of $\{ (x1,y1), (x2,y1) \}$. 
\vspace*{-1ex}
    \item We add an edge between two nodes in $G_j$ if some state variables corresponding to them are connected in the DBN through $\tta_j$. Formally, $E_j(u,v) \!=\! 1$,  
    if $\exists f,g \in F, \exists o_a\in O_a, j\in \{1,\ldots, |\A|\}$ s.t. the transition dynamics ($T^f$) for state variable $g'(o_v)$ and action $\tta_j(o_a)$ depend on state variable $f(o_u)$ or $f'(o_u)$.
    For the running example, there is no edge between $(x1,y1)$ and $(x2,y1)$ since cut-out, put-out or finisher's effects on one cell do not depend on any other cell.
\vspace*{-1ex}
    \item We add an edge between two nodes in $G_\K$  if some state variables corresponding to them are connected in the DBN (possibly through natural dynamics). I.e., $E_\K(u,v) \!=\! 1$,  
    if $\exists f,g \in F$ s.t. there is an edge from $f(o_u)$ (or $f'(o_u)$) and $g'(o_v)$ in the DBN.
    For the example, $E_4((x1,y1),(x2,y1)) = 1$ as there is an edge between $burning(x1,y1)$ and $burning'(x2,y1)$ since fire propagates to neighboring cells through natural dyanamics. 
    Similarly, $E_4((x2,y1),(x1,y1)) = 1$.
\vspace*{-1ex}
    \item As every node influences itself, self loops are added on each node. $E(v,v) = 1, \forall v \in V$. 
\end{enumerate}

For each node $v \in V$, we additionally construct a feature vector ($h(v)$) which consists of fluent feature vector ($h^f(v)$) and non-fluent feature vector ($h^{nf}(v)$), such that $h = concat (h^f, h^{nf})$. The feature vector for all nodes for the same object tuple is the same.
The feature vector is constructed as follows:
\begin{enumerate}
 \vspace*{-1ex}
   \item The fluent features for each node is obtained from the state of the problem instance. The values of state variables corresponding to a node are added as feature to that node. Whenever a fluent symbol cannot take a node as an argument, we add zero as the feature for it. Formally, $h^f(v)_i = g_i(o_v) ~\text{if}~ g_i \in \F, v \in V$ and $o_v$ is an argument of $g_i$, otherwise, $h^f(v)_i = 0, \forall i=1\ldots |\F|$. For the running example, we have two state-fluents. Hence, $h^f((x1,y1)) = [burning(x1,y1),out\mhyphen of\mhyphen fuel(x1,y1)]$.
\vspace*{-1ex}
    \item The non-fluent feature vector for each node is obtained from the RDDL file. The values of non-fluents defined on the node, and additionally any unary non-fluents where the argument intersects the node are added as the features for the node. 
    The default value is obtained from the domain file while the specific value (if available) is obtained from the instance file. Formally, $h^{nf}(v)_i = g_i(o_{nf}) ~\text{if}~ g_i \in \NF, v \in V, o_{nf} \in O_{nf}, ((o_v=o_{nf}) \lor (|o_{nf}|=1 \wedge o_{nf}\subset o_v))$, 
    otherwise, $h^f(v)_i = 0, \forall i=1\ldots |\NF|$.
    In our example, $h^{nf}((x1,y1)) = [target(x1,y1)]$. 
\vspace*{-1ex}
\end{enumerate}

We note that the size of feature vector on each node depends on the domain, but is independent of the number of objects in the instance -- there are a constant number of feature values per state predicate symbol. This allows variable-sized instances of the same domain to use the same representation.

\begin{table*}[t]\centering
\caption{$\alpha_{symnet}(0)$ values of \symnet{}. Bold values represent over $90\%$ the score of max performance.} 
\label{comparison_table}

\begin{tabular}{c @{\hspace{\tabcolsep}} ccccccc}
\toprule
& Instance & 5               & 6            & 7            & 8               & 9               & 10              \\
\midrule
\multirow{9}{*}{\rotatebox{90}{Domain}}
& AA       & \textbf{0.93 $\pm$ 0.01} & \textbf{0.94 $\pm$ 0.01} & \textbf{0.94 $\pm$ 0.01} & \textbf{0.92 $\pm$ 0.02} & \textbf{0.95 $\pm$ 0.03} & \textbf{0.91 $\pm$ 0.05} \\
& CT       & 0.87 $\pm$ 0.16          & 0.78 $\pm$ 0.14          & \textbf{1.00 $\pm$ 0.07} & \textbf{0.98 $\pm$ 0.13} & \textbf{0.99 $\pm$ 0.04} & \textbf{1.00 $\pm$ 0.05} \\
& GOL      & \textbf{0.96 $\pm$ 0.06} & \textbf{1.00 $\pm$ 0.05} & 0.65 $\pm$ 0.05          & 0.83 $\pm$ 0.03          & \textbf{0.95 $\pm$ 0.04} & 0.64 $\pm$ 0.08          \\
& Nav      & \textbf{0.99 $\pm$ 0.01} & \textbf{1.00 $\pm$ 0.01} & \textbf{0.99 $\pm$ 0.01} & \textbf{1.00 $\pm$ 0.01} & \textbf{1.00 $\pm$ 0.02} & \textbf{1.00 $\pm$ 0.02} \\
& ST       & \textbf{0.91 $\pm$ 0.05} & 0.84 $\pm$ 0.02          & 0.86 $\pm$ 0.05          & 0.85 $\pm$ 0.05          & 0.81 $\pm$ 0.02          & 0.89 $\pm$ 0.03          \\
& Sys      & \textbf{0.96 $\pm$ 0.03} & \textbf{0.98 $\pm$ 0.02} & \textbf{0.98 $\pm$ 0.02} & \textbf{0.97 $\pm$ 0.02} & \textbf{0.99 $\pm$ 0.01} & \textbf{0.96 $\pm$ 0.03} \\
& Tam      & \textbf{0.92 $\pm$ 0.07} & \textbf{1.00 $\pm$ 0.12} & \textbf{0.98 $\pm$ 0.06} & \textbf{1.00 $\pm$ 0.12} & \textbf{1.00 $\pm$ 0.12} & \textbf{0.95 $\pm$ 0.06} \\
& Tra      & 0.85 $\pm$ 0.18          & \textbf{0.93 $\pm$ 0.06} & 0.88 $\pm$ 0.21          & 0.74 $\pm$ 0.17          & \textbf{0.94 $\pm$ 0.12} & 0.87 $\pm$ 0.13          \\
& Wild     & \textbf{0.99 $\pm$ 0.01} & \textbf{1.00 $\pm$ 0.00} & \textbf{1.00 $\pm$ 0.00} & \textbf{1.00 $\pm$ 0.00} & \textbf{1.00 $\pm$ 0.01} & \textbf{1.00 $\pm$ 0.01} \\

\bottomrule
\end{tabular}
\end{table*}

\subsection{Representation Learning}
\symnet{} runs GAT on the instance graph to obtain node embeddings $\overline{v}$ for each node $v\in V$,  It then constructs tuple embedding for each object tuple by concatenating node embeddings of all associated nodes. Formally, let $O_V=\{o_v|v\in V\}$. For $o\in O_V$, the tuple embedding $\overline{o} = \mathrm{concat}(\overline{v})$, over all $v$ s.t. $o_v=o$. \symnet{} also computes a state embedding $\overline{s}$ by taking a dimension-wise max over all tuple embeddings, i.e., $\overline{s}=MaxPool_{o\in O_V}(\overline{o})$.



\subsection{Policy Decoder}

\symnet{} maps latent representations $\overline{o}$ and $\overline{s}$ into a state value $V(s)$ (long-term expected discounted reward starting in state $s$) and mixed policy $\pi(s)$ (probability distribution over all ground actions). This is done using a value decoder and a policy decoder, respectively. 

There are several challenges in designing a (generalized) policy decoder. First, the action symbols may take multiple objects as arguments. Second, and more importantly, action symbols may even take those object tuples as arguments that do not correspond to any node in the instance graph. This will happen if an object tuple (in $O_a$) is not an argument to any fluent symbol, i.e., $\exists o_a$ s.t. $o_a\in O_a \wedge o_a \notin O_f$. We note that adding these object tuples as nodes in the instance graph may not work, since we will not have any natural features for those nodes.


In response, we design a novel framework for policy and value decoders. The decoders consist of fully connected layers, the input to which are a subset of the tuple embeddings $\overline{o}$. \symnet{} uses the following rules to construct decoders:
\begin{enumerate}
 \vspace*{-1ex}
    \item  The number of decoders is constant for a given domain and is equal to the number of distinct action symbols ($|\A|$). For the running example, three different decoders for each policy and value decoding are constructed, namely $cut\mhyphen out, put\mhyphen out$ and $finisher$. 
\vspace*{-1ex}
    \item  The input to a decoder is the state embedding $\overline{s}$ concatenated with embeddings of object tuples corresponding to the state variables affected by the action in the DBN. In running example, $put\mhyphen out(x1,y1)$ action takes only the tuple embedding of $(x1,y1)$ as input. However, the number of state-variables being affected by a ground action might vary across instances of the same domain. For example, the $finisher$ action affects all cells. To alleviate this, we use size-independent max pool aggregation over the embeddings of all affected tuple embeddings to create a fixed-sized input. 
\vspace*{-1ex}
    \item Decoder parameters are specific to action symbols and not to ground actions. 
    In running example, $put\mhyphen out(x1,y1)$ will be scored using embedding of $(x1,y1)$; similarly, for $(x2, y1)$. But, both scorings will use a single parameter set specific to $put\mhyphen out$.
\vspace*{-1ex}
\item The policy decoder computes scores of all ground actions, which are normalized using softmax to output the final policy in a state. For $I_t$, the highest probability action is selected as the final action.
\vspace*{-1ex}
    \item All value outputs are summed to give the final value for that state. This modeling choice reflects the additive reward aspect of many RDDL domains.
\vspace*{-1ex}
\end{enumerate}

\subsection{Learning} 
While construction of \symnet{} architecture is heavily dependent on the RDDL domain and instance files, actual training is done via model-free reinforcement learning approach of A3C \cite{a3c}. RL learns from interactions with environment -- \symnet{} simulates the environment using RDDL-specified dynamics. Use of model-based planning algorithms for this purpose is left as future work. 
We formulate training of \symnet{} as a multi-task learning problem (see Section \ref{formulation}), so that it generalizes well and does not overfit on any one problem instance. The parameters for the state encoder, policy decoder, and value decoder are learned using updates similar to that in A3C. \symnet{}'s loss function for the policy and value network is the same as that in the A3C paper (summed over the multi-task problem instances). 

As constructed, \symnet{}'s number of parameters is independent of the size of the problem instance. Hence, the same network can be used for problem instances of any size. After the learning is completed, the network represents a generalized policy (or value), since it can be directly used on a new problem instance to compute the policy in a single forward pass.

\section{Experiments}

\begin{table*}[t]
\centering
\caption{Comparison of \symnet{} against  \symnet{}-s (SYM) architecture trained from scratch and \torpido{} (TOR) architecture trained from scratch. We compare out-of-the-box \symnet{} to others after 12 hours of training. INF is used when SYM or TOR achieved minimum possible reward and hence \symnet{} was infinitely better.} 
\label{comparison__from_scratch}
\scalebox{1}{
\begin{tabular}{cccccccccccc}
\toprule
Domain &  SYM & TOR & Domain &  SYM & TOR & Domain &  SYM & TOR & Domain &  SYM & TOR  \\
\cmidrule(lr){1-3}\cmidrule(lr){4-6}\cmidrule(lr){7-9}\cmidrule(lr){10-12}
AA 5   & 1.09          & 0.99          & GOL 5  & \textbf{1.35}  & \textbf{1.49} & ST 5    & 1.11           & 0.94           & Tam 5  & \textbf{INF}   & \textbf{2.33}  \\
AA 6   & 1.78          & 0.95          & GOL 6  & \textbf{1.57}  & \textbf{1.69} & ST 6    & 1.21           & 0.90           & Tam 6  & \textbf{27.71} & \textbf{8.13}  \\
AA 7   & 1.21          & 0.98          & GOL 7  & 1.08           & 0.76          & ST 7    & 1.10           & 0.87           & Tam 7  & \textbf{17.81} & \textbf{4.83}  \\
AA 8   & 1.31          & 0.97          & GOL 8  & 2.22           & 0.87          & ST 8    & 1.14           & 0.90           & Tam 8  & \textbf{2.74}  & \textbf{15.56} \\
AA 9   & 1.39          & 0.95          & GOL 9  & \textbf{1.86}  & \textbf{1.31} & ST 9    & 1.13           & 0.81           & Tam 9  & \textbf{24.94} & \textbf{13.07} \\
AA 10  & 1.32          & 0.93          & GOL 10 & 1.25           & 0.68          & ST 10   & 1.30           & 0.95           & Tam 10 & \textbf{2.35}  & \textbf{7.99}  \\
\cmidrule(lr){1-3}\cmidrule(lr){4-6}\cmidrule(lr){7-9}\cmidrule(lr){10-12}
CT 5   & \textbf{1.34} & \textbf{1.39} & Nav 5  & \textbf{10.84} & \textbf{INF}  & Sys 5   & \textbf{1.03}  & \textbf{2.89}  & Tra 5  & 1.78           & 0.86           \\
CT 6   & \textbf{INF}  & \textbf{1.56} & Nav 6  & \textbf{INF}   & \textbf{INF}  & Sys 6   & \textbf{1.33}  & \textbf{1.20}  & Tra 6  & \textbf{1.56}  & \textbf{1.39}  \\
CT 7   & \textbf{1.13} & \textbf{1.12} & Nav 7  & \textbf{INF}   & \textbf{INF}  & Sys 7   & \textbf{1.56}  & \textbf{2.45}  & Tra 7  & \textbf{3.28}  & \textbf{1.13}  \\
CT 8   & \textbf{1.55} & \textbf{1.23} & Nav 8  & \textbf{INF}   & \textbf{INF}  & Sys 8   & \textbf{1.46}  & \textbf{1.60}  & Tra 8  & 1.13           & 0.81           \\
CT 9   & \textbf{1.35} & \textbf{1.16} & Nav 9  & \textbf{INF}   & \textbf{INF}  & Sys 9   & \textbf{1.38}  & \textbf{1.17}  & Tra 9  & \textbf{2.50}  & \textbf{1.08}  \\
CT 10  & \textbf{1.22} & \textbf{4.99} & Nav 10 & \textbf{INF}   & \textbf{INF}  & Sys 10  & \textbf{1.18}  & \textbf{1.50}  & Tra 10 & \textbf{1.53}  & \textbf{1.86}  \\
\midrule
Wild 5 & \textbf{1.03} & \textbf{1.13} & Wild 7 & \textbf{1.03}  & \textbf{1.13} & Wild 9  & \textbf{1.01}  & \textbf{13.14} &        &                &                \\
Wild 6 & \textbf{1.01} & \textbf{1.01} & Wild 8 & \textbf{1.00}  & \textbf{1.09} & Wild 10 & \textbf{34.80} & \textbf{11.19} &        &                &                \\

\bottomrule
\end{tabular}}
\end{table*}

\begin{table}
\centering
\caption{Comparison of \trapsnet{} with \symnet{} on three domains as published in \citep{trapsnet}. Label: AA - Academic Advising, GOL - Game Of Life, Sys - Sysadmin}
\label{comp_trapsnet}
\begin{tabular}{ c @{\hspace{0.8\tabcolsep}} ccccccc}
\toprule
&Instance & 5               & 6            & 7            & 8               & 9               & 10              \\

\midrule
\multirow{3}{*}{\rotatebox{90}{Domain}}
&AA       & \textbf{1.12} & \textbf{1.17} & \textbf{1.12} & \textbf{1.27} & \textbf{1.26} & \textbf{1.40} \\
&GOL      & 0.96          & \textbf{1.04} & 0.69          & \textbf{1.00} & 0.97          & \textbf{1.50} \\
&Sys      & \textbf{1.01} & \textbf{1.55} & \textbf{1.33} & \textbf{1.39} & \textbf{1.21} & \textbf{1.17} \\

\bottomrule
\end{tabular}
\end{table}

Our goal is to estimate the effectiveness of \symnet{} out-of-the-box policy for a new problem in a domain. Unfortunately, there are no available transfer algorithms for general RDDL RMDPs. So, we first compare it against a random policy,  because that is the best we can do currently with no time to train. To further understand the overall quality of the generalized policy, we also compare it against several neural models that train from scratch on the test instance. We also compare it against state-of-the-art online planner PROST \cite{prost}. 


\subsection{Domains and Experimental Setting}
We show all our results on nine RDDL domains used IPPC 2014: Academic Advising (AA), Crossing Traffic (CT), Game of Life (GOL), Navigation (NAV), Skill Teaching (ST), Sysadmin (Sys), Tamarisk (Tam), Traffic  (Tra), and Wildfire (Wild). We describe the domains, and the number of state fluents, state non-fluents, and action fluents in the supplementary material.
The RL agent is trained to learn the generalized policy on smaller sized instances. We use IPPC problem instances 1, 2, and 3 of each domain for the multi-task training of \symnet{} network. 
In the spirit of domain-independent planning, we use the \emph{same} hyperparameters for each domain. The embedding module for GAT uses a neighborhood of 1 and an output feature size of 6. We then use a fully connected layer of output 20 dimensions to get an embedding from each of the tuple embedding outputs by GAT. All layers use a leaky ReLU activation and a learning rate of $10^{-3}$. We train the network using RMSProp \cite{rms} on a single Nvidia K40 GPU. \symnet{} is trained for each domain for twelve hours (4 hours for each instance).

\subsection{Comparison Algorithms and Metrics}
As there does not exist any previous method for learning over Relational RDDL MDPs, we can only compare against a random policy. However, this experiment can only show the difference from a random policy, but cannot evaluate the overall goodness of the generalized policy. For that, we compare against several (potentially upper bound) policies that are not directly comparable to \symnet{} in their experimental settings. For our first such experiment, we use \torpido{} as the state-of-the-art deep reactive policy. Note that we do not use their transfer method, but train the network from scratch on the problem instance. This is because it can only transfer across equi-sized instances. Still, it is an upper bound as \torpido{} trained on the test instance is compared against \symnet{} trained on other smaller instances, but not the test instance. Similarly, we also compare against \symnet{} architecture itself, trained from scratch on the test instance (named \symnet-s). The main difference between \torpido{} and \symnet{} architectures is that \torpido{} has a much higher capacity since it models each ground action explicitly. On three domains where \trapsnet{} is applicable, we also compare against \trapsnet{} policies out of the box. Finally, we also compare against the state-of-the-art {\em online} planner, PROST. 

After training algorithm $alg$ for $t$ hours, we simulate its output policy 200 times (for $H$ steps each) from the start state.  We average the discounted rewards to estimate the expected long term discounted reward of that policy, denoted by $V_{alg}(t)$.  To be able to compare across domains and problems and reward ranges, we report a normalized metric $\alpha_{alg}(t) = \frac{V_{alg}(t)-V_{min}}{V_{max}-V_{min}}$ where $V_{min}$ and $V_{max}$ are the minimum, and the maximum expected discounted rewards obtained at any time by any of the four comparison algorithms on a given instance.  This number lies between 0 and 1, with 1 being the best-found reward, and 0 being the random policy's reward. All algorithms are trained independently 5 times and the average result is reported.
During training from scratch, all networks start with a random policy and hence have their $\alpha(0)$ values as $0$. However, that is not true for \symnet{} as it is pre-trained on the domain. To compare against other training approaches directly, we compute $\beta_{alg}(t) = \frac{\alpha_{symnet}(0)}{\alpha_{alg}(t)}$. A value higher than 1 suggests that \symnet{} out-of-the-box outperformed $alg$ trained for $t$ hours, and less than 1 implies \symnet{} performed worse.

\begin{table*}[t]
\centering
\caption{Comparison of PROST with \symnet{}. INF is used when PROST returned a policy equal to or worse than a random policy.} 
\label{Comp_prost}
\scalebox{1.0}{
\begin{tabular}{c @{\hspace{\tabcolsep}} cccccccccc}
\toprule
&Domain & AA            & CT   & GOL  & Nav             & ST   & Sys           & Tam  & Tra  & Wild          \\
\midrule
\multirow{6}{*}{\rotatebox{90}{Instance}} 
&5        & \textbf{2.13} & 0.76 & 0.48 & \textbf{1.16}   & 0.95 & \textbf{1.13} & 0.61 & 0.60 & \textbf{1.39} \\
&6        & \textbf{2.14} & 0.44 & 0.57 & \textbf{1.87}   & 0.86 & \textbf{1.24} & 0.96 & 0.65 & \textbf{INF}  \\
&7        & \textbf{2.18} & 0.62 & 0.33 & \textbf{6.42}   & 0.86 & \textbf{1.13} & 0.70 & 0.61 & \textbf{INF}  \\
&8        & \textbf{1.79} & 0.37 & 0.39 & \textbf{45.46}  & 0.90 & \textbf{1.50} & 0.79 & 0.51 & \textbf{INF}  \\
&9        & \textbf{1.46} & 0.74 & 0.44 & \textbf{101.23} & 0.78 & \textbf{1.21} & 0.83 & 0.75 & \textbf{INF}  \\
&10       & \textbf{1.46} & 0.37 & 0.30 & \textbf{INF}    & 0.93 & \textbf{1.42} & 0.84 & 0.64 & \textbf{1.49} \\

\bottomrule
\end{tabular}}
\end{table*}

\subsection{Results}

{\bf Comparison against Random Policy: } We report the values of $\alpha_{symnet}(0)$ in Table \ref{comparison_table}. Since the random policy is 0, we notice that on all six problem instances from the nine domains, \symnet{} performs enormously better than random. We highlight the instances where our method achieves over $90\%$ of the max reward obtained by any algorithm for that instance. We see that \symnet{} with no training achieves over $90\%$ the max reward on 40 instances and over $80\%$ in 50 out of 54 instances. We also show that our method performs the best out-of-the-box in 28 instances. This is our main result, and it highlights that \symnet{} takes a major leap towards the goal of computing generalized policies for the whole RMDP domain, and can work on a new instance out of the box.

\noindent{\bf Comparison against Training from Scratch: }  
We now compare \symnet{} against the expected discounted rewards obtained by \torpido{} and \symnet{}-s, when they are trained from scratch for 12 hours on the test problem. We note that these numbers are not directly comparable, since in one case, the model has been trained on other instances of the domain, but not trained on the test problem at all, and in the other case the models are trained from scratch on the test. That said, this comparison is likely a good indicator of the absolute performance of \symnet{}. 

Table \ref{comparison__from_scratch} reports the values for $\beta_{torpido}(12)$ and $\beta_{symnet-s}(12)$. We notice that, surprisingly, \symnet{} policy with no training is better than both methods on several instances. 
Against \symnet{} trained from scratch, it is better on all instances, although its edge over \torpido{} is limited to 37 out of 54. We hypothesize that this excellent performance is due to the multi-task learning aspect of \symnet{}, where it is able to reach some generalized policy of a domain that is not found on the specific instance even after training for 12 hours. 

In 17 out of 54 instances, \symnet{} lags behind \torpido, which is not surprising, since \torpido{} has much higher capacity, as discussed earlier. We also notice that the performance of \torpido{} is no better than random for Navigation. We attribute this to the sparse and late reward obtained in large instances of this domain, which makes it difficult for \torpido{} to learn a good policy. Because of the late rewards, \torpido{} is not able to reach the goal state at all in 12 hours of training, and hence is not able to improve on the random policy. \symnet{} trains well on small instances where path to goal is short and generalizes well. In GOL, \symnet{} performs worse, because the nature of policy changes significantly in large instances (e.g. requiring new patterns to survive) which cannot be learned in smaller instances at all.


\noindent{\bf Comparison against \trapsnet{}: } While \trapsnet{} is not applicable in many RMDPs, still, we can compare it with \symnet{} on some domains. We compare these on three domains that follow the unary fluents and binary non-fluents constraint: Academic Advising, Game of Life, and SysAdmin. We report $\beta_{trapsnet}(0)$ in Table \ref{comp_trapsnet}. It shows that \symnet{} outperforms \trapsnet{} on 15 out of 18 instances, is comparable on 2 instances and worse on 1 instance. We attribute the success of \symnet{} over \trapsnet{} to the action-symbol specific graphs ($G_j$),  which likely help learn better action dependencies in the embeddings.


\noindent{\bf Comparison against ASNets: }
Even after significant efforts, we were not able to compare against ASNets, which solves a similar problem for PPDDL domains.  
Converting an RDDL  domain to PPDDL enumerates all the ground state-variables and loses the RMDP structure. This leads to different domain files for different instances for the same problem domain, due to which ASNets is unable to train. We also tried writing a domain file manually for a few domains, but were not successful due to the unavailability of floating non-fluent values, and due to non-additive reward structure in PPDDL.


\noindent{\bf Comparison against PROST: }
Finally, we compare against PROST. PROST is a state-of-the-art {\em online} planner, i.e., it performs interleaved planning and execution, as it builds a new search tree before taking every action, based on the specific state reached. On the other hand, \symnet{} outputs an offline policy, which does not need much computation for deciding the next action. Offline and online policies are two very different settings, and these results are \emph{not directly comparable}. Nonetheless, we report $\frac{V_{symnet}(0)-V_{min}}{V_{prost}-V_{min}}$. The code of PROST is obtained from the official repository and we use its default settings for this comparison.\footnote{\url{https://github.com/prost-planner/prost}} 

We compare our policy with PROST on all 9 domains, shown in Table \ref{Comp_prost}. We see that on four domains \symnet{} achieves a much better performance than PROST. This is rather surprising to us that even after substantial lookahead from the current state, PROST is still not able to compute a good policy. For example, in both Navigation and Wildfire, the rewards are sparse and distant, and PROST is often unable to reach the goal in its planning horizon.
In other five domains, PROST is substantially better than \symnet.  This suggests that \symnet{} policies are not close to optimal, and further research is needed for making them even stronger. This also points to the future possibility of applying a combination of \symnet{} and PROST for the offline setting, not unlike the use of Monte-Carlo Tree Search with deep neural networks in AlphaGo \cite{alphago}.

Overall, we find that \symnet{}'s generalized policies out-of-the-box are enormously better than random, and can frequently beat other deep neural models trained from scratch on the test instance. However, comparison with PROST suggests that \symnet{} policies are not close to optimal and further research is needed to make them even better.
\section{Conclusion and Future Work}
We present the first neural-method for obtaining a generalized policy for Relational MDPs represented in RDDL. Our method, named \symnet{}, converts an RDDL problem instance into an instance graph, on which a graph neural network computes state embeddings and embeddings for important object tuples. These are then decoded into scores for each ground action. All parameters are tied and size-invariant such that the same model can work on problems of varying sizes. In our experiments, we train \symnet{} on small problems of a domain and test them on larger problems to find that they out-of-the-box perform hugely better than random. Even when compared against training deep reactive policies from scratch, \symnet{} without training perform better or at par in over half the problem instances. 

Our work is an attempt to revive the thread on Relational MDPs and the attractive vision of generalized policies for a domain. However, ours is only one of the first steps. Further investigation is needed to assess how far are \symnet{}'s generalized policies from optimal. We strongly believe that there may be even better architectures that could learn near-optimal generalized policies, and the need for retraining or interleaving planning and execution could be rendered unnecessary. We release all our software for use by the research community at \url{https://github.com/dair-iitd/symnet}.


\section*{Acknowledgements}

We would like to thank Scott Sanner for an extremely insightful discussion on DBNs, which enabled us to work out a solution to convert an RMDP into a graph using DBNs. We also thank Vishal Sharma, Gobind Singh, Parag Singla and the anonymous reviewers for their comments on various drafts of the paper. This work is supported by grants from Google, Bloomberg, IBM and 1MG, Jai Gupta Chair Fellowship, and a Visvesvaraya faculty award by Govt. of India. We thank Microsoft Azure sponsorships, and the IIT Delhi HPC facility for computational resources.

\bibliography{refs}
\bibliographystyle{icml2020}

\clearpage

\appendix
\section*{Appendix}
\section{Domain Description}
We describe the details of the domains presented in the IPPC 2011 and IPPC 2014. The statistics for state fluents ($\F$), non-fluents ($\NF$) and Action ($\A$) for all the domains are show in the Table \ref{domain_action_stat} and Table \ref{domain_state_stat}. UP represent $\F$, $\NF$ and $\A$ without parameters, Unary represents $\F$, $\NF$ and $\A$ with a single parameter and multiple represents $\F$, $\NF$ and $\A$ with more than one parameter. Table \ref{onj_list} lists the instance specific number of objects, state variables and action variables for the domain. The domains $1,2,3$ are used for training, $4$ for validation and $5,6,7,8,9,10$ for testing.

\subsection*{Academic Advising}
The academic advising domain represents a student at a university trying to complete his/her degree. Some courses are required to be completed to obtain the final degree. Each course is either a basic course or may have prerequisites. The probability of passing a course depends on the number of prerequisites completed (a fixed probability if no prerequisite). The goal is to complete the degree as soon as possible.

\subsection*{Crossing Traffic}
Crossing Traffic is represents a robot in a grid, with obstacles at a random grid cell at any time. The obstacles (car) start at any cell randomly and move left. The robot aims to plan its path from the starting grid cell to the goal cell while avoiding obstacles.

\subsection*{Game of Life}
Game of Life domain is represented as a grid where each cell can either be dead or alive. The goal is to keep as many cells alive as possible. The probability of cell death depends on the number of neighbors alive at a particular time, which is non-linear in the number of neighbors alive.

\begin{table}[t]
\centering

\caption{The statistics related to the domains listing the number of UP (Un-Paramataried), Unary and Muiltiple Action ($\A$) for each domain.}
\label{domain_action_stat}
\begin{tabular}{ccccccc}
\toprule 
Domain & UP-$\A$ & Unary-$\A$ & Multiple-$\A$  \\
\midrule
Academic Advising & 0 & 1 & 0 \\
Crossing Traffic & 4 & 0 & 0\\
Game of Life & 0 & 0 & 1 \\
Navigation & 4 & 0 & 0 \\
Skill Teaching & 0 & 2 & 0 \\
Sysadmin & 0 & 1 & 0\\
Tamarisk & 0 & 2 & 0 \\
Traffic & 0 & 1 & 0 \\
Wildfire & 0 & 0 & 2 \\
\bottomrule
\end{tabular}
\end{table}

\begin{table*}[t]
\centering
\caption{The statistics related to the domains listing the number of UP (Un-Paramataried), Unary and Multiple State Fluents ($\F$) and Non-Fluents ($\NF$) for each domain.}
\label{domain_state_stat}
\begin{tabular}{ccccccc}
\toprule 
Domain & UP-$\F$ & UP-$\NF$ & Unary-$\F$ & Unary-$\NF$ & Multiple-$\F$ & Multiple-$\NF$ \\
\midrule
Academic Advising & 0 & 1 & 2 & 5 & 0 & 1\\
Crossing Traffic & 0 & 1 & 0 & 4 & 2 & 5\\
Game of Life & 0 & 0 & 0 & 0 & 1 & 2\\
Navigation & 0 & 0 & 0 & 4 & 1 & 6\\
Skill Teaching & 0 & 0 & 6 & 7 & 0 & 1\\
Sysadmin & 0 & 2 & 1 & 0 & 0 & 1\\
Tamarisk & 0 & 17 & 2 & 0 & 0 & 2\\
Traffic & 0 & 0 & 3 & 3 & 0 & 3\\
Wildfire & 0 & 4 & 0 & 0 & 2 & 2\\
\bottomrule
\end{tabular}
\end{table*}

\subsection*{Navigation}
Navigation represents a robot in a grid world where the aim is to reach a goal cell as quickly as possible. The probability of the robot dying in a particular cell is different, which is specified in the instance file.

\subsection*{Skill Teaching}
Skill Teaching domain represents a teacher trying to teach a skill to students. Each student has a mastery level in a particular skill. Some skills have pre-conditions, which increase the probability of learning a particular skill. The skill is taught using either hints or multiple-choice questions. The goal is to answer as many questions as possible by the student by learning the required skill.

\subsection*{Sysadmin}
Sysadmin domain represents computers connected in a network. The probability of a computer shutting down on its own depends on the number of turned-on neighboring computers. The agent can either turn on a computer or leave it as it is. The goal is to maximize the number of computers at a particular time.

\subsection*{Tamarisk}
Tamarisk domain represents invasive species of plants (Tamarisk) trying to take over native plant species. The plants spread in any direction and try to destroy the native plant species. The agent can either eradicate Tamarisk in a cell or restore the native plant species, each having a different reward. The goal is to minimize the cost of eradication and restoration of the native plant species.

\subsection*{Traffic}
Traffic domain models the traffic on the road with roads connecting at various intersections. Each road intersection has two traffic light signals combinations of which yield different traffic movement. The agent aims to control the traffic signal (only on the forward sequence) to control the traffic.

\subsection*{Wildfire}
The wildfire domain represents a forest catching fire. The direction of fire spreading depends on the direction of the wind and also the type of fuel at that point (e.g., grass or wood, etc.). The agent can either choose to put down the fire or cut off the fuel even before the fire happens. The goal is to prevent as many cells as possible, and more reward is provided to protect high priority cells.

\section{Variation of $\alpha_{\symnet{}}(0)$ with neighbourhood}

\begin{figure*}
    \centering
    \subfloat[AA]{\includegraphics[scale=0.33]{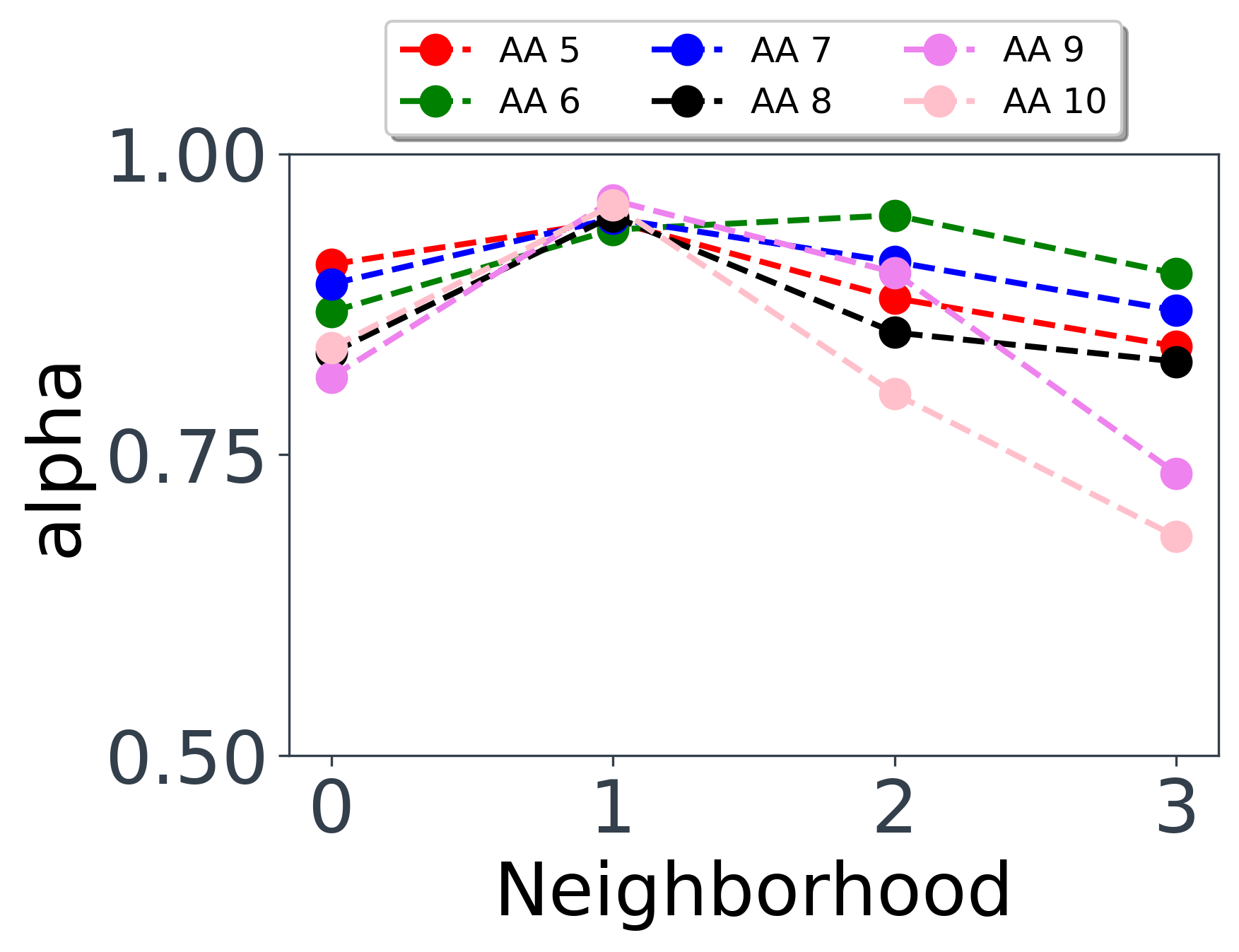}}
    \subfloat[CT]{\includegraphics[scale=0.33]{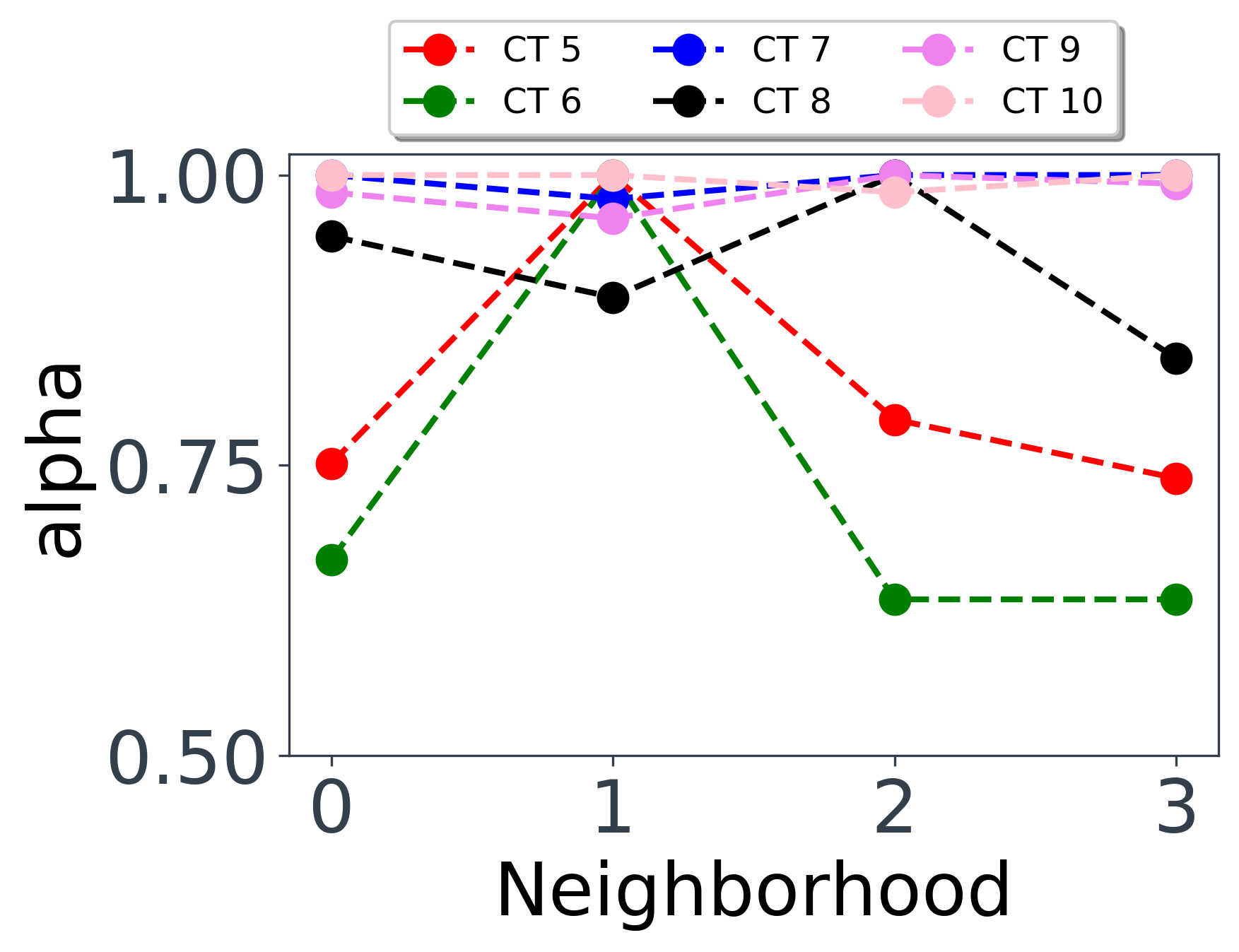}}
    \subfloat[GOL]{\includegraphics[scale=0.33]{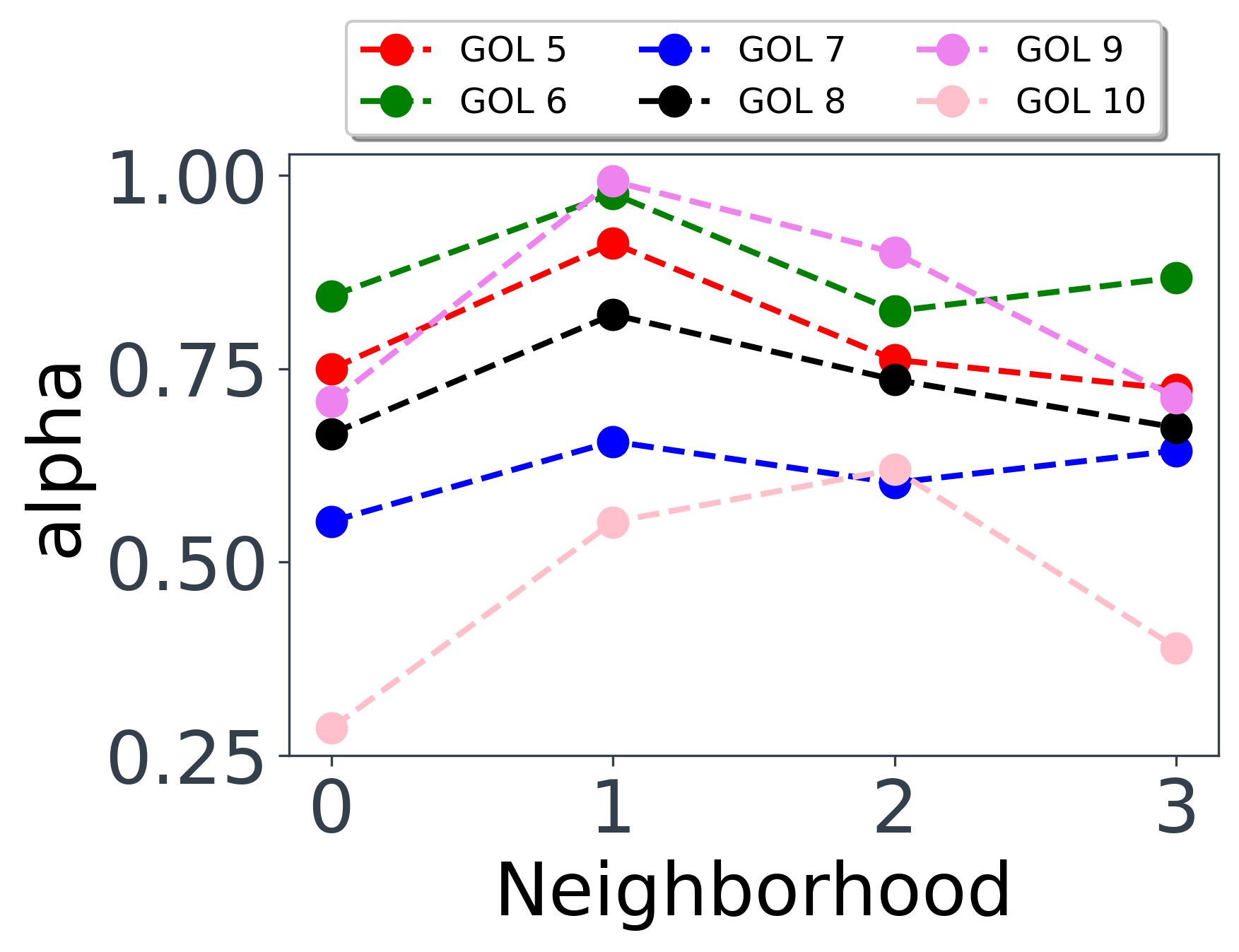}}\\
    \subfloat[Nav]{\includegraphics[scale=0.33]{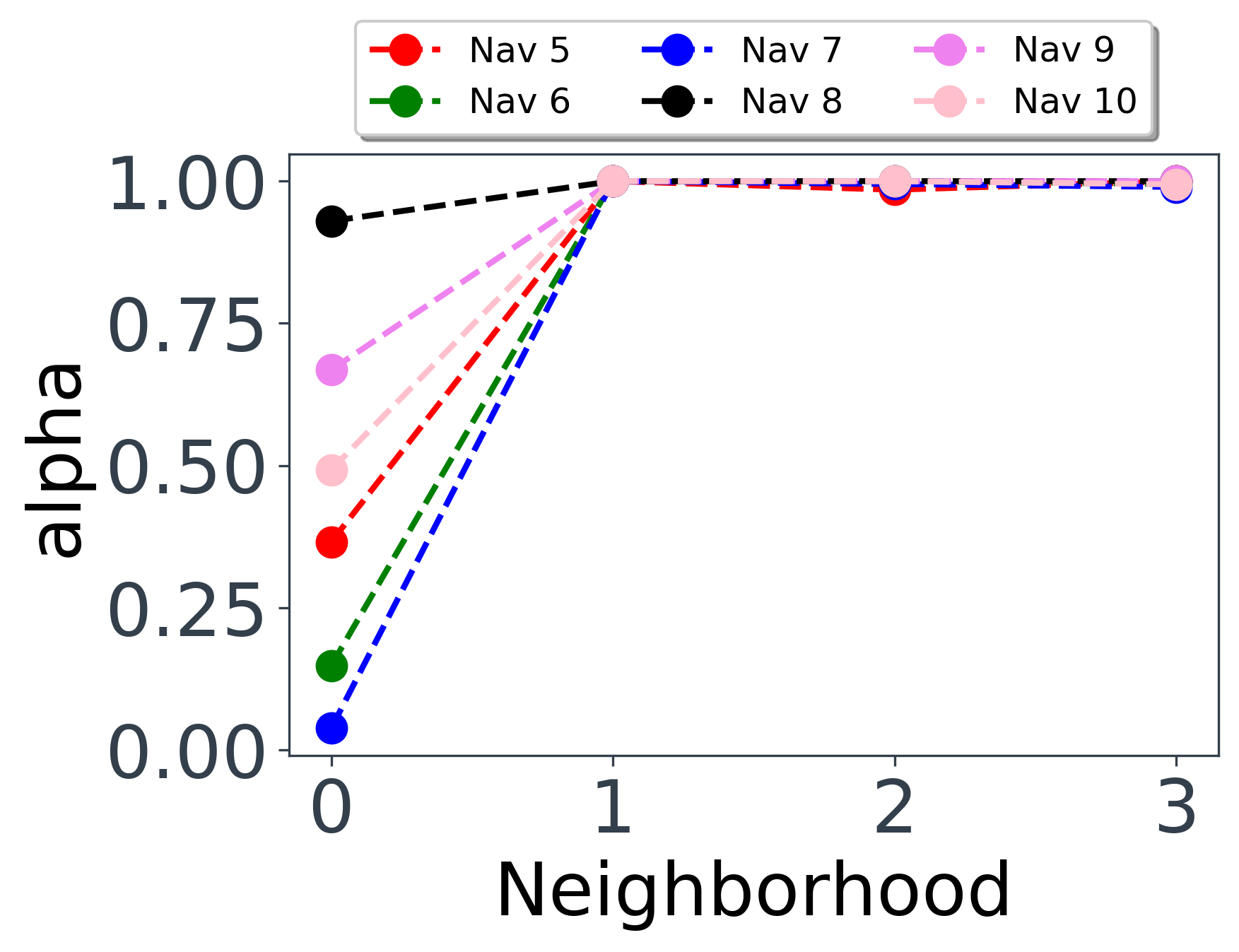}}
    \subfloat[ST]{\includegraphics[scale=0.33]{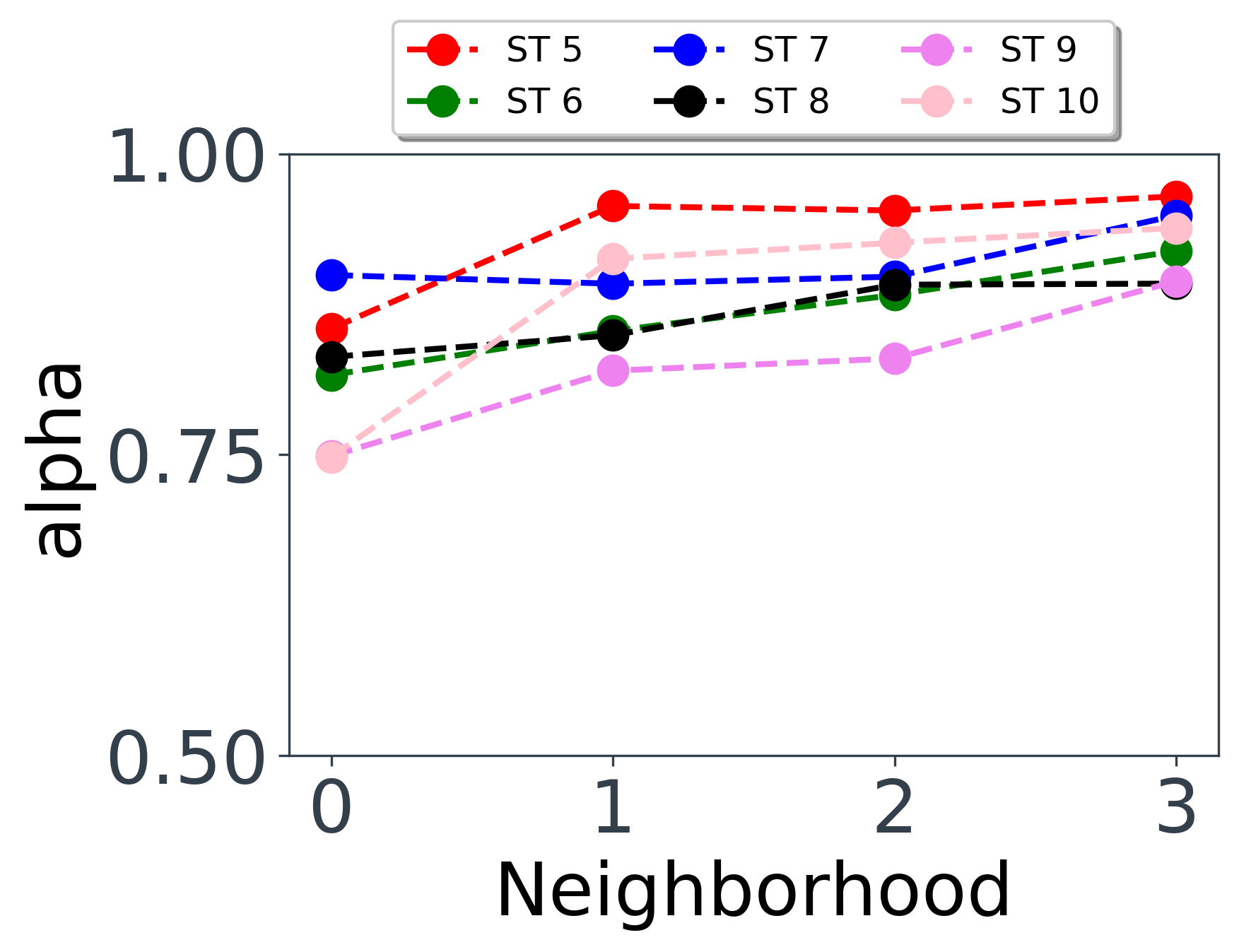}}
    \subfloat[Sys]{\includegraphics[scale=0.33]{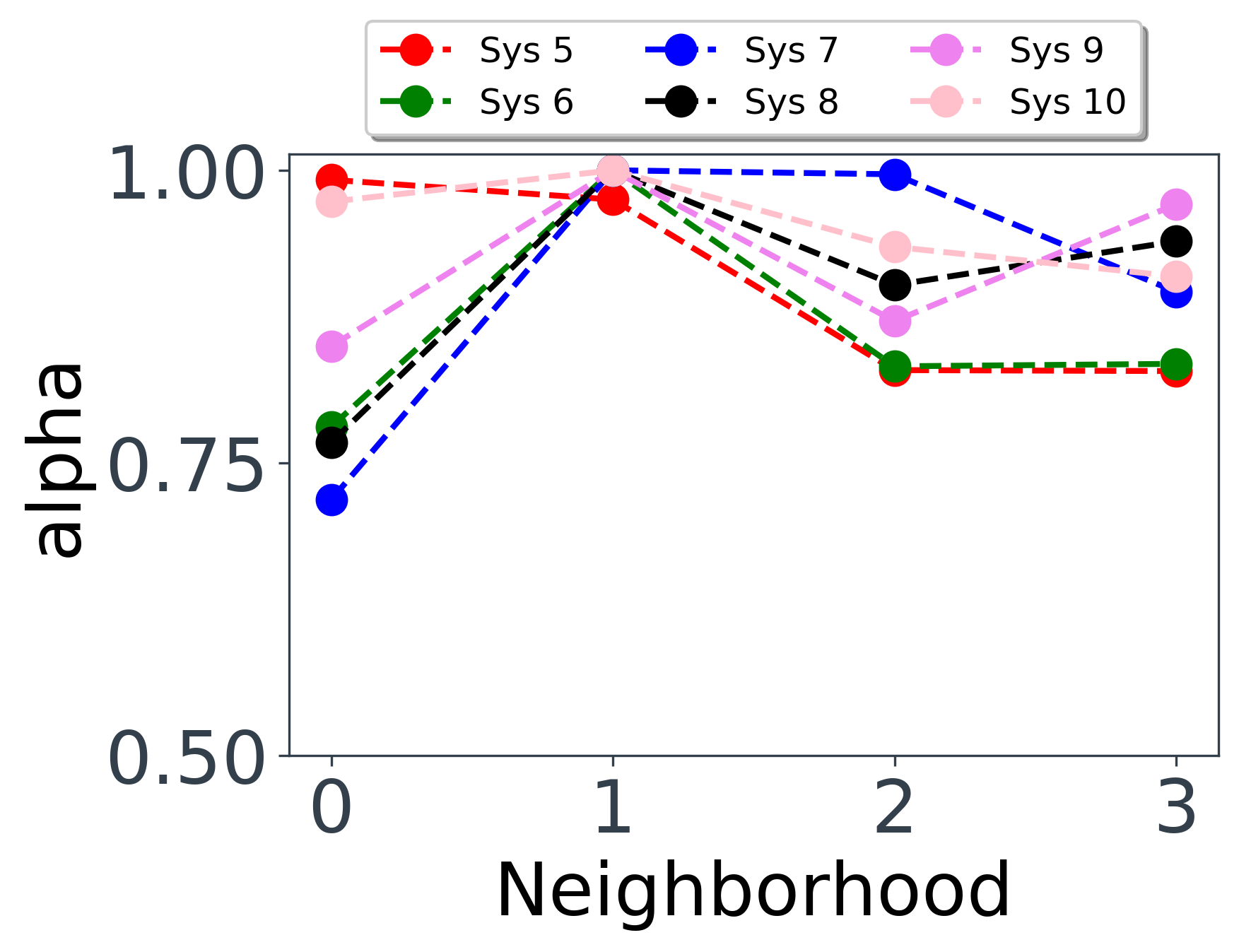}}\\
    \subfloat[Tam]{\includegraphics[scale=0.33]{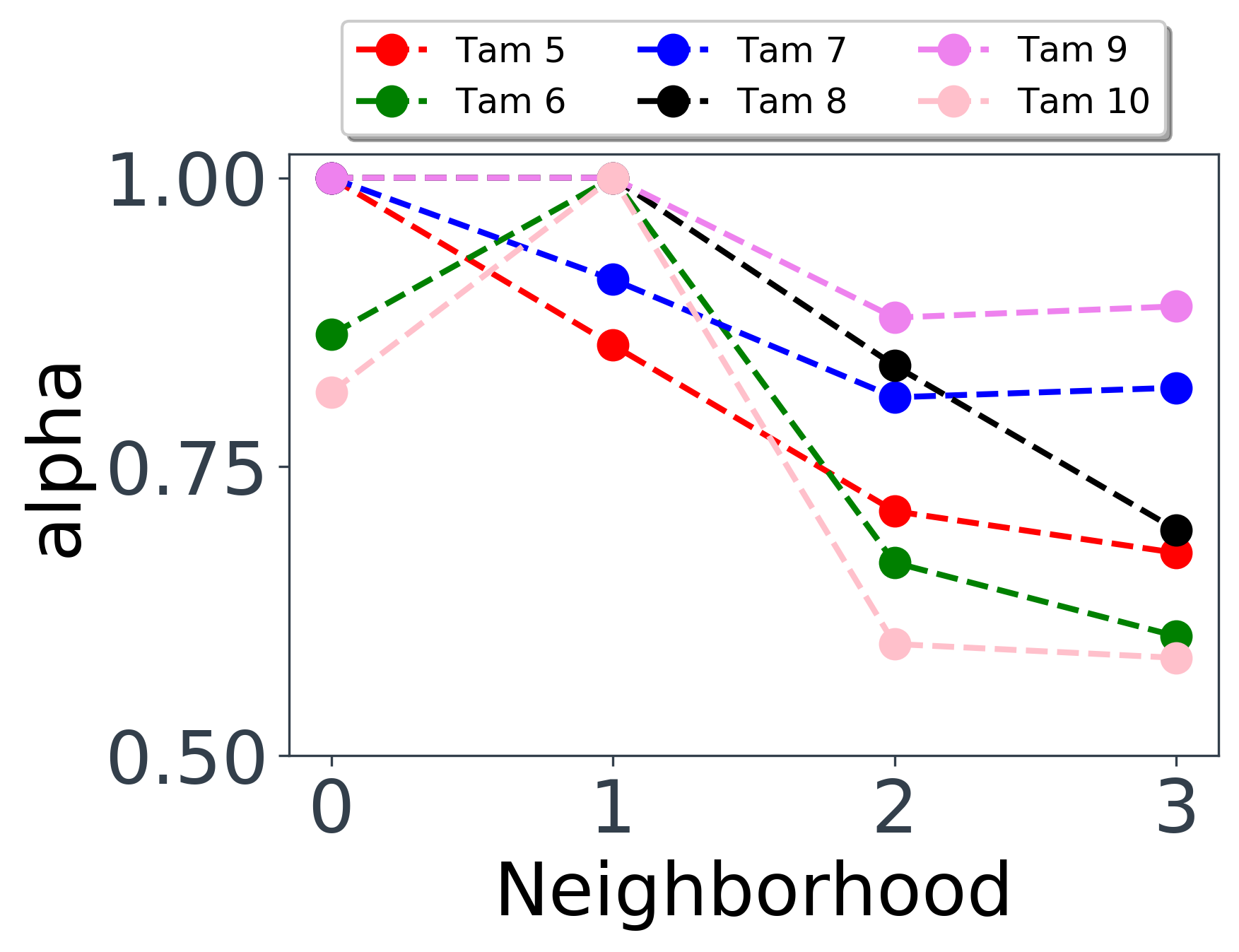}}
    \subfloat[Tra]{\includegraphics[scale=0.33]{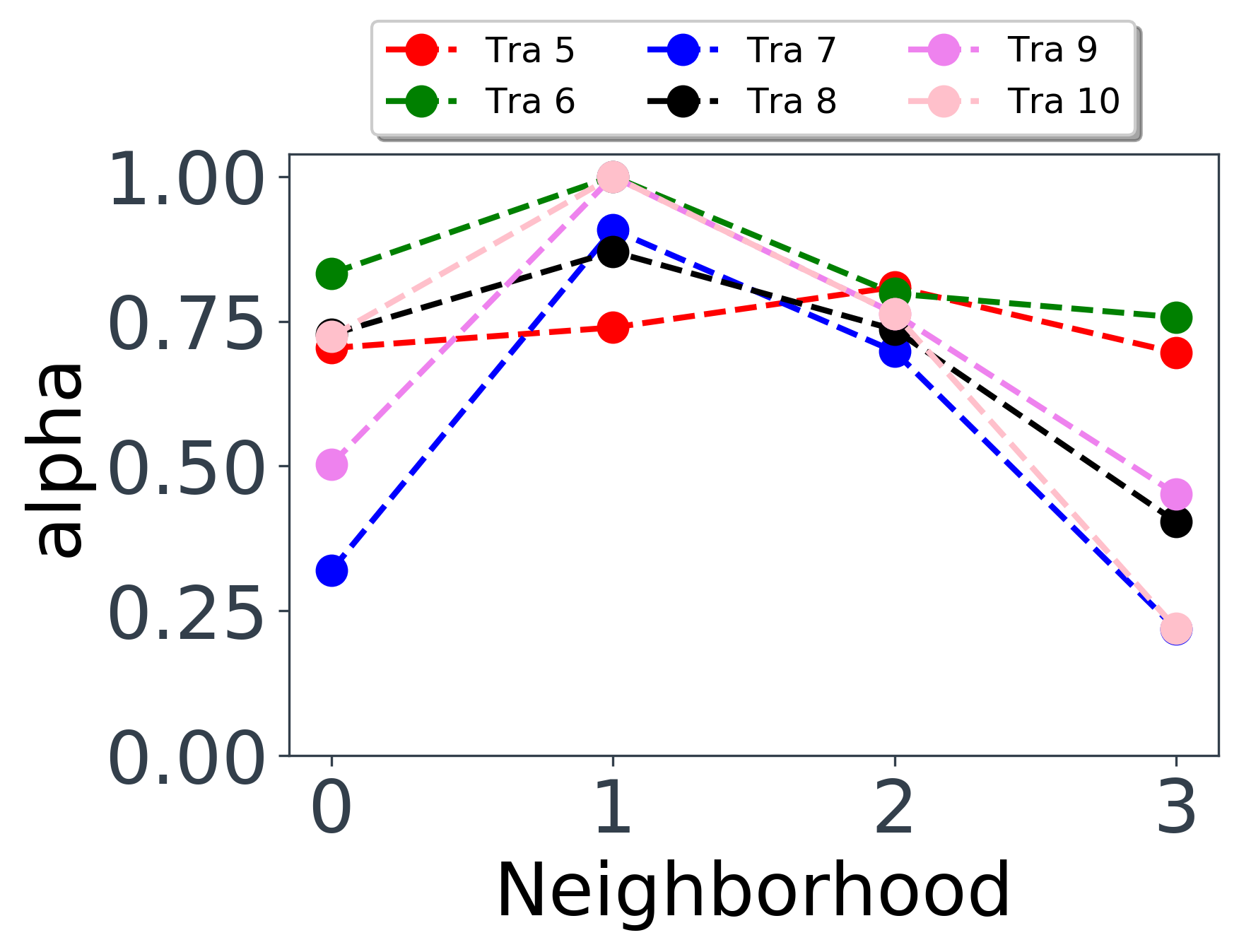}}
    \subfloat[Wild]{\includegraphics[scale=0.33]{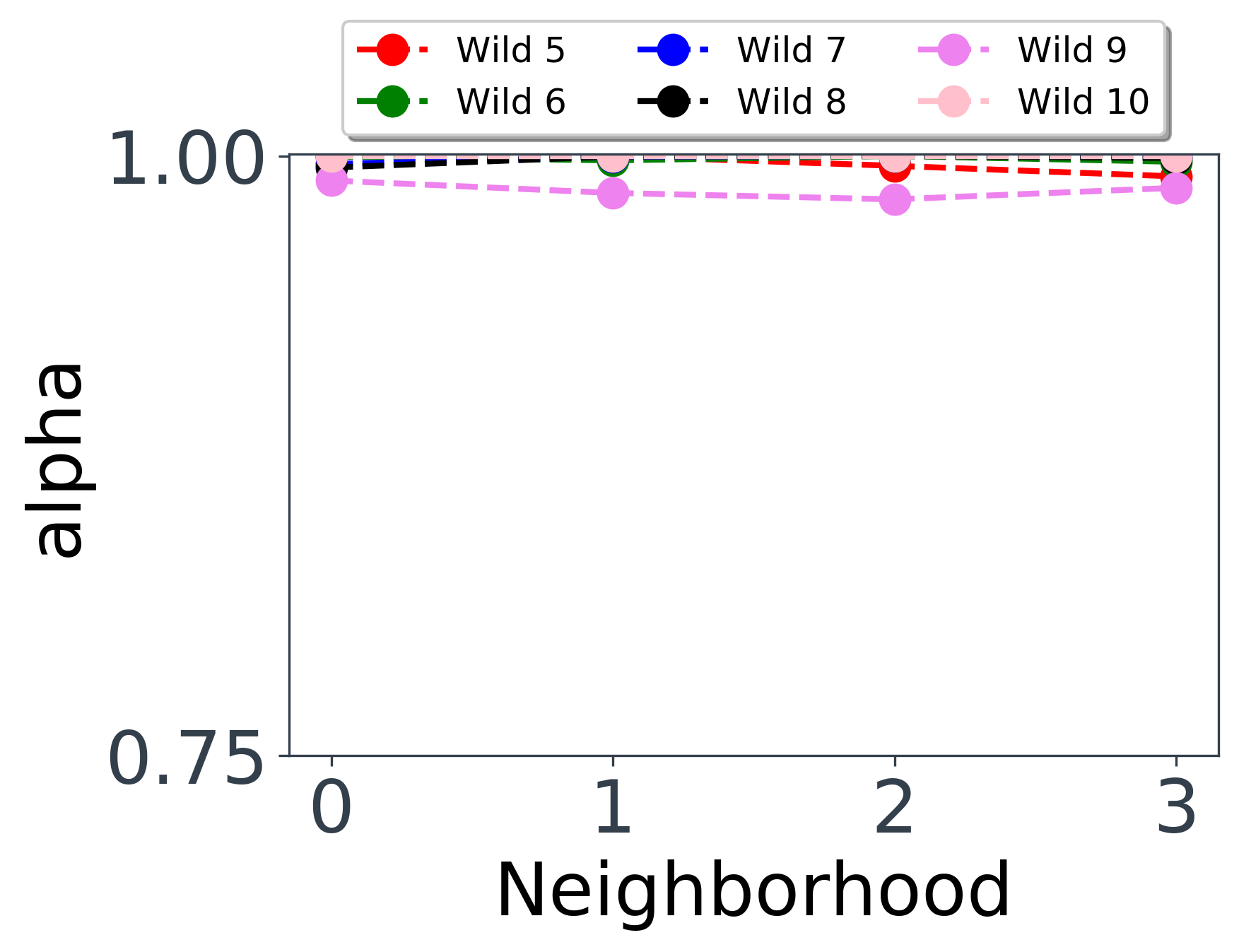}}
    
    \caption{Variation of $\alpha_{\symnet{}}(0)$ with neighbourhood. [Larger is better]}
    \label{alpha_variation}
\end{figure*}

To inspect the importance of the neighborhood information in learning a generalized policy for the domains, we perform the study of the neighborhood parameter variation. In the Figure \ref{alpha_variation}, we show the variation of $\alpha_{\symnet{}}(0)$ with neighbourhood. From the Figure, we observe that message passing for the neighborhood of size $1$ yields the best results for most domains, and hence we reported the results with neighborhood $1$ in the main paper. In general, we observe that the value of $\alpha_{\symnet{}}(0)$ first increases and then decreases.\par 
For most instances, the $\alpha_{\symnet{}}(0)$ is less for neighborhood $0$ compared to neighborhood $1$, showing that the information regarding the neighbors is necessary for learning a better policy. For example, in domain academic advising, the neighborhood $1$ aggregates information about the pre-requisites for the courses and then prioritizes the courses to take. A similar trend is observed in domain skill teaching, where the information about the pre-condition for the skill plays an important role in learning the skills. For some domains like navigation, neighborhood information is absolutely critical for planning the next move which can be observed from very low values of $\alpha_{\symnet{}}(0)$ from Figure \ref{alpha_variation}(d). Other domains like wildfire are not affected a lot by neighborhood a lot. This is because the margin between the minimum and maximum rewards is large, and the generalized policy outputs rewards close to the maximum value, which decreases the variation in the value of $\alpha_{\symnet{}}(0)$. As we increase the value of neighborhood to $2$ and $3$, the value of $\alpha_{\symnet{}}(0)$ tends to fall down for most instances. We hypothesize that the agent overfits to instance-specific policies for the instances it is trained on and hence fails to generalize.

\begin{table*}[t]
\centering
\caption{The statistics related to the domain instances listing the number of Objects, State Variables and Action Variables for all the instances of the domains. Domain $1,2,3$ are used for training, $4$ for validation and $5,6,7,8,9,10$ for testing.}
\label{onj_list}
\begin{tabular}{cccccccc}
\toprule
Domain & \#Objects & \#State Vars & \#Action Vars & Domain  & \#Objects & \#State Vars & \#Action Vars \\
\cmidrule(lr){1-4}\cmidrule(lr){5-8}
AA 1   & 10        & 20           & 11            & ST 1    & 2         & 12           & 5             \\
AA 2   & 10        & 20           & 11            & ST 2    & 2         & 12           & 5             \\
AA 3   & 15        & 30           & 16            & ST 3    & 4         & 24           & 9             \\
AA 4   & 15        & 30           & 16            & ST 4    & 4         & 24           & 9             \\
AA 5   & 20        & 40           & 21            & ST 5    & 6         & 36           & 13            \\
AA 6   & 20        & 40           & 21            & ST 6    & 6         & 36           & 13            \\
AA 7   & 25        & 50           & 26            & ST 7    & 7         & 42           & 15            \\
AA 8   & 25        & 50           & 26            & ST 8    & 7         & 42           & 15            \\
AA 9   & 30        & 60           & 31            & ST 9    & 8         & 48           & 17            \\
AA 10  & 30        & 60           & 31            & ST 10   & 8         & 48           & 17            \\
\cmidrule(lr){1-4}\cmidrule(lr){5-8}
CT 1   & 9         & 12           & 5             & Sys 1   & 10        & 10           & 11            \\
CT 2   & 9         & 12           & 5             & Sys 2   & 10        & 10           & 11            \\
CT 3   & 16        & 24           & 5             & Sys 3   & 20        & 20           & 21            \\
CT 4   & 16        & 24           & 5             & Sys 4   & 20        & 20           & 21            \\
CT 5   & 25        & 40           & 5             & Sys 5   & 30        & 30           & 31            \\
CT 6   & 25        & 40           & 5             & Sys 6   & 30        & 30           & 31            \\
CT 7   & 36        & 60           & 5             & Sys 7   & 40        & 40           & 41            \\
CT 8   & 36        & 60           & 5             & Sys 8   & 40        & 40           & 41            \\
CT 9   & 49        & 84           & 5             & Sys 9   & 50        & 50           & 51            \\
CT 10  & 49        & 84           & 5             & Sys 10  & 50        & 50           & 51            \\
\cmidrule(lr){1-4}\cmidrule(lr){5-8}
GOL 1  & 9         & 9            & 10            & Tam 1   & 12        & 16           & 9             \\
GOL 2  & 9         & 9            & 10            & Tam 2   & 16        & 24           & 9             \\
GOL 3  & 9         & 9            & 10            & Tam 3   & 15        & 20           & 11            \\
GOL 4  & 16        & 16           & 17            & Tam 4   & 20        & 30           & 11            \\
GOL 5  & 16        & 16           & 17            & Tam 5   & 18        & 24           & 13            \\
GOL 6  & 16        & 16           & 17            & Tam 6   & 24        & 36           & 13            \\
GOL 7  & 25        & 25           & 26            & Tam 7   & 21        & 28           & 15            \\
GOL 8  & 25        & 25           & 26            & Tam 8   & 28        & 42           & 15            \\
GOL 9  & 25        & 25           & 26            & Tam 9   & 24        & 32           & 17            \\
GOL 10 & 30        & 30           & 31            & Tam 10  & 32        & 48           & 17            \\
\cmidrule(lr){1-4}\cmidrule(lr){5-8}
Nav 1  & 12        & 12           & 5             & Tra 1   & 28        & 32           & 5             \\
Nav 2  & 15        & 15           & 5             & Tra 2   & 28        & 32           & 5             \\
Nav 3  & 20        & 20           & 5             & Tra 3   & 40        & 44           & 5             \\
Nav 4  & 30        & 30           & 5             & Tra 4   & 40        & 44           & 5             \\
Nav 5  & 30        & 30           & 5             & Tra 5   & 52        & 56           & 5             \\
Nav 6  & 40        & 40           & 5             & Tra 6   & 52        & 56           & 5             \\
Nav 7  & 50        & 50           & 5             & Tra 7   & 64        & 68           & 5             \\
Nav 8  & 60        & 60           & 5             & Tra 8   & 64        & 68           & 5             \\
Nav 9  & 80        & 80           & 5             & Tra 9   & 76        & 80           & 5             \\
Nav 10 & 100       & 100          & 5             & Tra 10  & 76        & 80           & 5             \\
\midrule
Wild 1 & 9         & 18           & 19            & Wild 6  & 25        & 50           & 51            \\
Wild 2 & 9         & 18           & 19            & Wild 7  & 30        & 60           & 61            \\
Wild 3 & 16        & 32           & 33            & Wild 8  & 30        & 60           & 61            \\
Wild 4 & 16        & 32           & 33            & Wild 9  & 36        & 72           & 73            \\
Wild 5 & 25        & 50           & 51            & Wild 10 & 36        & 72           & 73          \\
\bottomrule
\end{tabular}
\end{table*}
\clearpage

\end{document}